\documentclass[letterpaper, 10 pt, conference]{ieeeconf}  

\IEEEoverridecommandlockouts                              

\usepackage{hyperref}
\usepackage{graphics} 
\usepackage{epsfig} 
\usepackage{mathptmx} 
\usepackage{times} 
\usepackage{amsmath} 
\usepackage{amssymb}  
\usepackage{hyphenat} 

\usepackage{adjustbox}
\usepackage{booktabs}
\usepackage{colortbl}
\usepackage[table]{xcolor}
\usepackage{tabularx}
\usepackage{makecell}
\usepackage{longtable}
\usepackage{microtype}
\usepackage{threeparttable}
\usepackage{threeparttablex}
\usepackage{pifont} 
\newcommand{\cmark}{\ding{51}}
\newcommand{\xmark}{\ding{55}}
\usepackage{booktabs}
\usepackage{multirow}
\newcommand{\NA}{--}
\newcommand{\method}{PV-WM}
\definecolor{dgray}{gray}{0.45}

\newcolumntype{Q}[1]{>{\centering\arraybackslash}m{#1}}
\definecolor{dgray}{gray}{0.45}
\providecommand{\sdev}[1]{{\scriptsize$\pm$#1}}
\providecommand{\gcell}[1]{\textcolor{dgray}{#1}}
\providecommand{\NA}{\textit{n/a}}
\providecommand{\abpm}[2]{#1{\scriptsize$\pm$#2}}
\newcommand{\dcaeobb}{\ensuremath{\mathrm{DCAE}_{\mathrm{OBB}}}}

\title{\LARGE \bf
PV-WM: A Heterogeneous Micro--Macro World Model for Articulated Pedestrian--Vehicle Co-Rollout
}

\author{Haozhuang Chi$^{1}$, Jingsong Liang$^{1}$, Ziying Song$^{1}$,
Lei Yang$^{1}$, Shihao Li$^{2}$, Haoruo Zhang$^{1}$, and
Chen Lv$^{1,\dagger}$%
\thanks{$^{1}$Nanyang Technological University, Singapore.}%
\thanks{$^{2}$Beijing Institute of Technology, Beijing, China.}%
\thanks{$^{\dagger}$Corresponding author, email:
{\tt\small lyuchen@ntu.edu.sg}.}%
}

\begin{document}

\maketitle
\thispagestyle{empty}
\pagestyle{empty}

\begin{abstract}

Local pedestrian--vehicle forecasting spans heterogeneous physical scales: pedestrians combine root locomotion with articulated motion, whereas vehicles are rigid bodies described by kinematic state and oriented extent. Existing road-agent forecasters typically omit pedestrian articulation, while pose forecasters leave vehicle futures outside the learned rollout. We introduce \method{}, a history-only world model over structured post-perception tracks. It recurrently advances pedestrian root motion, 15-joint articulation, and learned vehicle states within a synchronized heterogeneous state. The generated pedestrian and vehicle chunks jointly form the next recurrent boundary; vehicle boxes are reconstructed from predicted
center and heading with observed extent, and P--V geometry is recomputed after every transition. Relative to a matched one-shot complete-state predictor, recurrent execution reduces Root ADE by \(12.7\%\) and MPJPE by \(14.8\%\). Feedback interventions show that later predictions depend on the content, temporal order, and pedestrian identity of generated articulation. Across 824 aligned Waymo contexts, with 797 providing valid future vehicle support, \method{} reduces Root ADE by \(5.2\%\), MPJPE by \(7.6\%\), P--V distance error by \(11.9\%\), and oriented-box closest-approach error by \(5.8\%\) relative to a validation-selected Modular Specialist. The single-network model uses \(57.1\%\) fewer parameters, \(96.5\%\) lower average FLOPs per local scene, and \(25.5\%\) lower measured p95 latency. PV-WM unifies this heterogeneous future state while preserving type-specific pedestrian and vehicle dynamics.

\end{abstract}

\begin{figure*}[!t]
    \centering
    \includegraphics[width=0.98\linewidth]{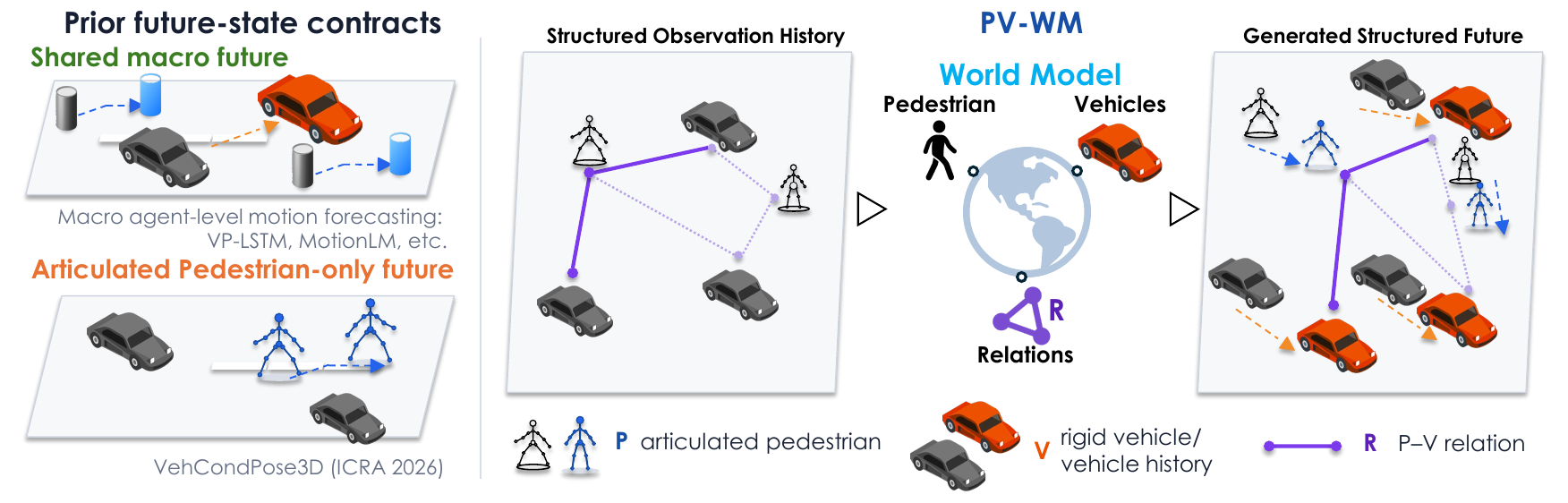}
    \caption{\textbf{Motivation for heterogeneous co-rollout.}
    \textit{Left}: Agent-level forecasting omits pedestrian articulation,
    whereas vehicle-conditioned pose forecasting leaves vehicle futures
    outside the rollout. \textit{Right}: PV-WM advances articulated
    pedestrians and learned rigid-vehicle states together. Generated P/V chunks jointly form the next boundary, P--V geometry is recomputed, and the persistent pair state is updated for the next transition.}
    \label{fig:teaser}
\end{figure*}

\section{Introduction}
\label{sec:introduction}

The state and transition define the predictive scope of a world model.
Local pedestrian--vehicle scenes are heterogeneous across physical type
and resolution: pedestrians combine root transport with articulated body
motion, vehicles carry rigid kinematic state and oriented extent, and
P--V geometry evolves with both agents' motion. Point- or
box-level pedestrian states omit body configuration, whereas body-only
forecasting leaves the surrounding vehicle future unresolved. Because
these variables evolve on the same clock, a complete local model must
advance them within one synchronized future.

Existing forecasting paradigms model different subsets of this state. Multi-agent forecasters predict trajectories, centers, boxes,
or motion tokens
~\cite{womd,trajectronpp,scene_transformer,wayformer,mtr,qcnet,motionlm,gameformer},
but generally retain pedestrians at agent-level resolution.
Articulated forecasters preserve global locomotion and local pose
~\cite{tbiformer,t2p,waymo3dskelmo}, while vehicle-conditioned pose
models use traffic histories without generating vehicle futures
~\cite{vehcondpose3d}. Driving world models span agent, occupancy,
latent, visual, and driver-centered state spaces
~\cite{trafficbots,occworld,driveworld,drivewm,epona,chi2026driverwm},
but do not expose articulated pedestrian state, learned vehicle motion,
oriented extent, and synchronized P--V geometry within one recurrent
local state. Composing separate pose and vehicle predictors can fill
the same output fields, but does not define a shared transition in
which both generated states form the next boundary and their geometry
is resynchronized after every step.

We introduce \textbf{\method{}}, a history-only heterogeneous
micro--macro world model for articulated pedestrian--vehicle co-rollout
over structured post-perception tracks. It predicts pedestrian root
motion and 15-joint articulation together with learned vehicle states
over five recurrent chunks. The generated P/V chunks jointly form the two-frame boundary for the
next transition; vehicle boxes are reconstructed and P--V geometry is recomputed after every transition.
PV-WM thus evolves its own structured state across the horizon rather
than decoding the future once from a fixed history representation.

State completeness and pair-to-endpoint routing are independent design
choices. PV-WM combines type-specific endpoint transitions with a
persistent pair state while keeping the endpoint branches factorized.
Routing controls inject pair context into the pedestrian branch, the
vehicle branch, or both.

Ablations attribute the principal architectural gains to recurrent typed
execution and the persistent pair branch, while relational supervision
improves closest-approach accuracy. Feedback interventions show that
later transitions reuse generated articulation. Across a segment-grouped, context-disjoint five-fold evaluation of 824 Waymo contexts, including 797 with valid
vehicle and pair support, \method{} improves Root ADE, MPJPE, P--V
distance error, and \dcaeobb{} over a validation-selected Modular
Specialist by \(5.2\%\), \(7.6\%\), \(11.9\%\), and \(5.8\%\).
The single-network model uses \(57.1\%\) fewer parameters,
\(96.5\%\) lower average FLOPs per local scene, and \(25.5\%\)
lower measured p95 latency.

Our contributions are threefold:
\begin{itemize}
    \item We formulate pedestrian--vehicle forecasting as recurrent
    prediction of a heterogeneous state spanning pedestrian transport
    and articulation, learned vehicle dynamics, oriented extent, and
    synchronized pair geometry.

    \item We introduce a generated-state co-rollout with typed endpoint
    dynamics and persistent pair state, and isolate the effect of
    pair-to-endpoint routing under matched conditions.

    \item We isolate the contributions of recurrent execution, typed
    dynamics, and the persistent pair branch, establish functional reuse of
    generated articulation, and demonstrate a more accurate and efficient
    complete-state system than a validation-selected Modular Specialist.
\end{itemize}

\section{Related Work}
\label{sec:related_work}

\subsection{Multi-Agent Motion Forecasting}
\label{sec:rw_motion_forecasting}

Multi-agent motion forecasting predicts road-user futures from agent
histories and scene context. The Waymo Open Motion Dataset formalizes
marginal and joint forecasting~\cite{womd}. Trajectron++ combines
heterogeneous dynamics with semantic context~\cite{trajectronpp},
while scene- and query-centric architectures reason over agents, time,
intentions, and maps through attention
~\cite{scene_transformer,wayformer,mtr,qcnet}. M2I decomposes
interactive prediction into influencer and conditional-reactor
forecasts~\cite{m2i}, whereas FJMP factorizes joint futures over a
learned directed acyclic interaction graph~\cite{fjmp}. MotionLM
autoregressively generates multi-agent futures as motion tokens and
supports temporally causal conditional rollout~\cite{motionlm};
GameFormer uses hierarchical game-theoretic refinement
~\cite{gameformer}. VP-LSTM directly combines pedestrian trajectories
with vehicle oriented-box forecasting~\cite{vplstm}.

These approaches span marginal, conditional, joint, autoregressive, and
graph-factorized prediction, but retain pedestrians at agent-level
resolution. PV-WM instead carries articulated pedestrian state
alongside learned vehicle motion on a shared recurrent timeline.

\subsection{Articulated Human Motion Forecasting}
\label{sec:rw_articulated_motion}

Articulated forecasting preserves the distinction between global
locomotion and root-relative body motion. TBIFormer models inter- and
intra-person dynamics with trajectory-aware body-part tokens
~\cite{tbiformer}, while T2P conditions local pose on predicted global
trajectories~\cite{t2p}. Waymo-3DSkelMo brings temporally coherent 3D
skeletons into autonomous-driving scenes~\cite{waymo3dskelmo}.
VehCondPose3D further conditions future pedestrian pose on surrounding
vehicle histories through vehicle encoding and pedestrian--vehicle
attention~\cite{vehcondpose3d}. These methods preserve body
configuration, but surrounding vehicles remain observed conditions
rather than generated future states. PV-WM advances articulated
pedestrians and learned vehicles within the same future process.

\subsection{World Models for Autonomous Driving}
\label{sec:rw_world_models}

World models learn predictive transitions for imagination, simulation,
and decision making~\cite{worldmodels,dreamerv3}. Driving world models
instantiate these transitions at different resolutions: TrafficBots
rolls forward explicit multi-agent behavior~\cite{trafficbots},
OccWorld predicts future occupancy and ego motion~\cite{occworld}, and
DriveWorld learns latent scene dynamics for downstream perception and
planning~\cite{driveworld}. Visual models such as GAIA-1, Drive-WM,
and Epona generate future observations from scene, trajectory, or
action conditions~\cite{gaia1,drivewm,epona}. Driver-WM recurrently
forecasts articulated in-cabin motion conditioned on traffic context~\cite{chi2026driverwm}.

PV-WM targets a complementary external post-perception state in which
pedestrian articulation, root transport, learned rigid-vehicle state,
oriented extent, and P--V geometry remain explicit and synchronized
throughout rollout.

\section{Method}
\label{sec:method}

\begin{figure*}[!t]
    \centering
    \includegraphics[width=0.98\linewidth]{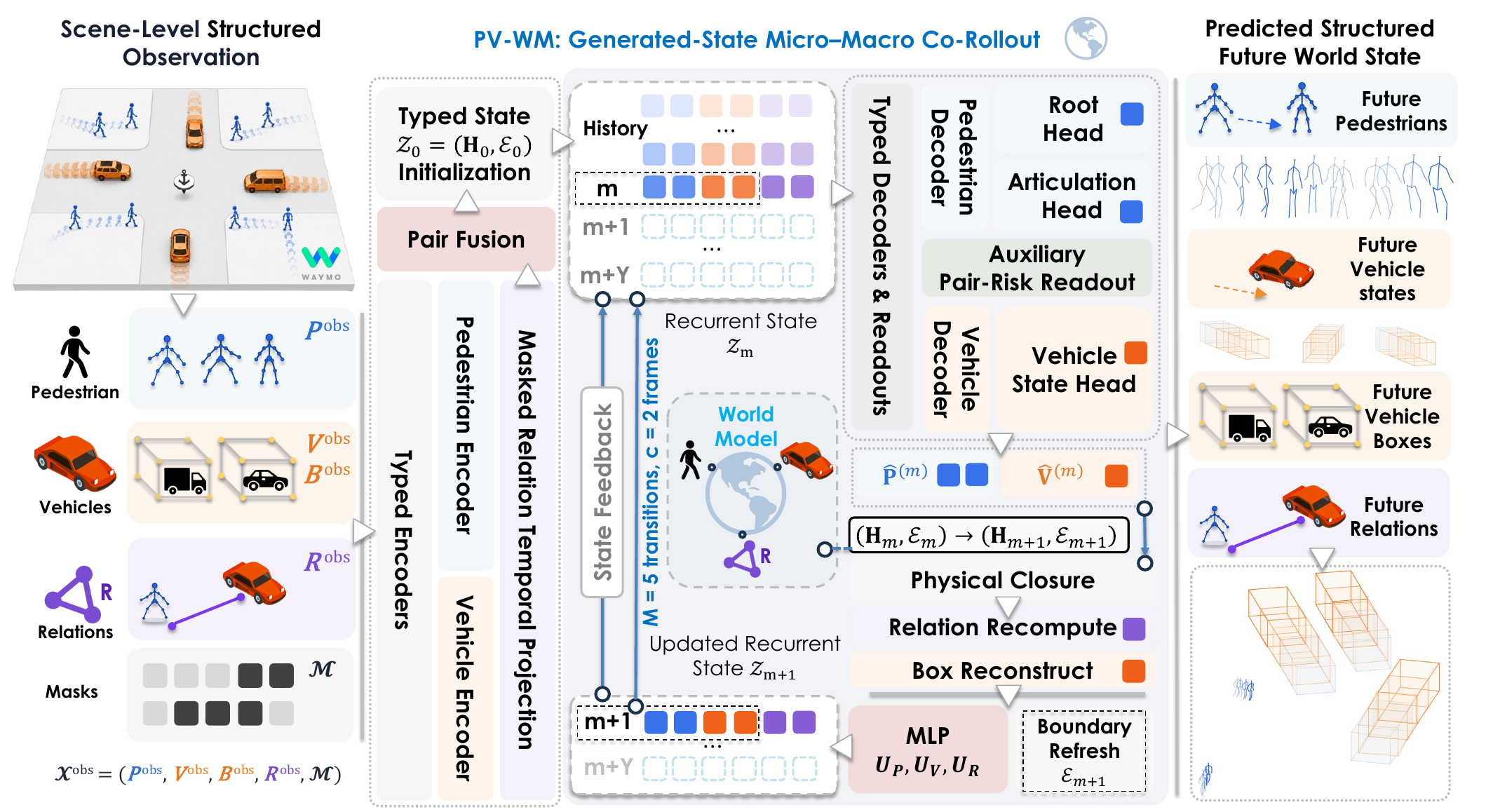}
    \caption{\textbf{PV-WM architecture.}
    Typed encoders and the final two observed P/V states initialize
    \(\mathcal Z_0=(\mathbf H_0,\mathcal E_0)\). Each transition decodes
    P/V chunks, recomputes relation geometry, reconstructs oriented boxes,
    updates typed memories, and refreshes \(\mathcal E_{m+1}\). Five
    transitions produce the structured future.}
    \label{fig:method}
\end{figure*}

\method{} advances structured post-perception tracks through a recurrent
rollout. At the macro scale, it predicts pedestrian root motion and
rigid-vehicle kinematics; at the micro scale, it predicts root-relative
pedestrian articulation. All variables evolve on a shared clock. The
model observes \(T_o=20\) frames and predicts \(T_f=10\) frames at
\(10\,\mathrm{Hz}\) through \(M=5\) transitions of \(c=2\) frames.
Generated P/V chunks jointly form the next two-frame boundary; vehicle
OBBs are reconstructed and P--V geometry is recomputed from
the generated states after every transition. PV-WM combines typed endpoint dynamics with a
persistent pair state while keeping the endpoint branches factorized.
Routing controls inject pair context into the pedestrian branch, the
vehicle branch, or both.

\subsection{Heterogeneous State and Typed Encoding}
\label{sec:method_state_encoding}

Indices \(i\), \(k\), and \(a\in\{0,\ldots,14\}\) denote pedestrians,
vehicles, and joints, with \(a=0\) the pelvis root; \(t\), \(m\), and
\(\tau\) denote frame, transition, and within-chunk time. Valid components
use \(\mathcal M=(\boldsymbol\mu^P,\boldsymbol\mu^V,\mathbf M^R)\), with
\(\mu^R_{ik}=\max_t M^R_{ik,t}\). We translate coordinates by the final
observed mean pedestrian root without rotating the scene.

The typed states are
\begin{equation}
\begin{aligned}
\mathbf p_{i,t,a}
&=\mathbf p_{i,t,0}+\boldsymbol\ell_{i,t,a},
&\boldsymbol\ell_{i,t,0}&=\mathbf 0,\\
\mathbf s^V_{k,t}
&=(\mathbf x^V_{k,t},\mathbf u^V_{k,t},
\sin\psi_{k,t},\cos\psi_{k,t},\nu_{k,t},o_{k,t}).
\end{aligned}
\label{eq:typed_state}
\end{equation}
Here \(\mathbf p_{i,t,0}\) is root transport,
\(\boldsymbol\ell_{i,t,a}\) is root-relative articulation, and
\(\mathbf x^P_{i,t}=\Pi_{xy}(\mathbf p_{i,t,0})\).
For vehicles, \(\mathbf x^V_{k,t}\), \(\mathbf u^V_{k,t}\),
\(\psi_{k,t}\), \(\nu_{k,t}\), and \(o_{k,t}\) denote planar center,
planar velocity, heading, speed, and presence, respectively. Let
\(\mathbf P_t=\{\mathbf p_{i,t,a}\}_{i,a}\) and
\(\mathbf V_t=\{\mathbf s^V_{k,t}\}_k\). For each valid pair,
\begin{align}
\boldsymbol\delta_{ik,t}
&=\mathbf x^V_{k,t}-\mathbf x^P_{i,t},
&d_{ik,t}&=\|\boldsymbol\delta_{ik,t}\|_2,\nonumber\\
\kappa_{ik,t}
&=-\frac{\boldsymbol\delta_{ik,t}^{\top}
(\mathbf u^V_{k,t}-\mathbf u^P_{i,t})}{d_{ik,t}+\epsilon},
&\mathbf r_{ik,t}&=(\delta^x,\delta^y,d,\kappa)_{ik,t},
\label{eq:relation_state}
\end{align}
where \(\mathbf u^P\) is a finite difference and positive \(\kappa\)
denotes closing motion. With
\(\mathbf R_t=\{\mathbf r_{ik,t}\}_{i,k}\) and \(\mathbf B_t\) the
vehicle OBB collection, the explicit state is
\(\mathcal S_t^{\mathrm{phys}}=(\mathbf P_t,\mathbf V_t,\mathbf R_t,\mathbf B_t)\),
where \(\mathbf R_t=\Phi(\mathbf P_{t-1:t},\mathbf V_t)\) and
\(\mathbf B_t=\Psi_{\mathrm{box}}(\mathbf V_t;
\boldsymbol\xi^{V,\mathrm{obs}})\). The structured input and physical-state output are
\begin{equation}
\begin{aligned}
\mathcal X^{\mathrm{obs}}
&=(\mathbf P^{\mathrm{obs}},\mathbf V^{\mathrm{obs}},
\mathbf B^{\mathrm{obs}},\mathbf R^{\mathrm{obs}},\mathcal M),\\
\hat{\mathcal S}^{\mathrm{phys}}
&=(\hat{\mathbf P},\hat{\mathbf V},\hat{\mathbf R},\hat{\mathbf B}).
\end{aligned}
\label{eq:state_contract}
\end{equation}
Only P/V states are generated directly. \(\Phi\) applies
Eq.~\eqref{eq:relation_state}; \(\Psi_{\mathrm{box}}\) uses generated
center and heading with the observed vertical center and static extent
\(\boldsymbol\xi^{V,\mathrm{obs}}\), normalizing heading only for box
construction. \(\hat{\mathbf Q}\) denotes an auxiliary pair-risk
readout optimized through \(\mathcal L_Q\).

The pedestrian encoder aggregates root, anatomical-group, and
root-relative motion; the vehicle GRU pools center increments, velocity,
acceleration, heading, speed, and presence. It initializes
\begin{equation}
\begin{aligned}
\tilde{\mathbf h}^P_i&=E_P(\mathbf P_i^{\mathrm{obs}}),
&\tilde{\mathbf h}^V_k&=E_V(\mathbf V_k^{\mathrm{obs}}),\\
\mathbf h^R_{ik,0}
&=F_R[\tilde{\mathbf h}^P_i\Vert\tilde{\mathbf h}^V_k\Vert
E_R(\mathbf R^{\mathrm{obs}}_{ik};\mathbf M^R_{ik})].
\end{aligned}
\label{eq:pair_memory_initialization}
\end{equation}
Endpoint adapters initialize \(\mathbf h^P_{i,0}\) and
\(\mathbf h^V_{k,0}\) from their corresponding typed encodings. All
routing settings initialize and retain \(\mathbf h^R_{ik,0}\). Routing
controls additionally inject mask-aggregated pair context into selected
endpoint branches during initialization, decoding, and recurrent
updates.

\subsection{Generated-State Co-Rollout}
\label{sec:method_rollout}

At chunk boundary \(m\in\{0,\ldots,M\}\),
\begin{equation}
\begin{aligned}
\mathcal Z_m&=(\mathbf H_m,\mathcal E_m),\\
\mathbf H_m&=(\{\mathbf h^P_{i,m}\}_i,\{\mathbf h^V_{k,m}\}_k,
\{\mathbf h^R_{ik,m}\}_{i,k}),\\
\mathcal E_m&=(\mathbf P_m^{\mathrm{prev}},\mathbf P_m^{\mathrm{cur}},
\mathbf V_m^{\mathrm{prev}},\mathbf V_m^{\mathrm{cur}}),\\
\mathcal E_0&=\operatorname{Tail}_2(\mathbf P^{\mathrm{obs}},
\mathbf V^{\mathrm{obs}}),
&\mathcal Z_0&=(\mathbf H_0,\mathcal E_0).
\end{aligned}
\label{eq:recurrent_world_state}
\end{equation}
The encoders compress the full \(T_o\)-frame history into \(\mathbf H_0\).
At each boundary, \(\mathbf H_m\) carries typed context, including the
persistent pair state, while \(\mathcal E_m\) supplies adjacent P/V
states for constant-velocity references and finite-difference geometry.
The explicit relation \(\mathbf R_t\) is recomputed from P/V states,
whereas \(\mathbf h^R_{ik,m}\) retains pair history. Parameters
\(\theta\) are shared across all \(M\) transitions; for \(m\geq1\),
\(\mathcal E_m\) contains generated P/V states only. For
\(m=0,\ldots,M-1\), the transition is
\begin{equation}
\begin{aligned}
&(\hat{\mathcal S}^{\mathrm{phys},(m)},\hat{\mathbf Q}^{(m)},
\mathcal Z_{m+1})\\
&\qquad=\mathcal T_{\theta,\boldsymbol{\rho}}
(\mathcal Z_m;\mathcal M,
\boldsymbol\xi^{V,\mathrm{obs}}),\quad
\boldsymbol{\rho}=(\rho_P,\rho_V)\in\{0,1\}^2.
\end{aligned}
\label{eq:world_transition}
\end{equation}
Here \(\rho_P\) and \(\rho_V\) gate pair context to the pedestrian and
vehicle endpoint branches, respectively. \(\mathcal M\) and
\(\boldsymbol\xi^{V,\mathrm{obs}}\) remain fixed across rollout. Root and vehicle motion use constant-velocity residuals;
articulation uses a gated residual from the current root-relative pose:
\begin{align}
\hat{\mathbf p}_{i,m,\tau,0}
&=\mathbf p^{\mathrm{cv}}_{i,m,\tau,0}+\Delta\mathbf p^P_{i,m,\tau},
\nonumber\\
\hat{\boldsymbol\ell}_{i,m,\tau,a}
&=\boldsymbol\ell^{\mathrm{cur}}_{i,m,a}
+\sigma(\gamma_{i,m,\tau,a})\Delta\boldsymbol\ell_{i,m,\tau,a},
\quad a=1,\ldots,14,\nonumber\\
\hat{\mathbf s}^{V}_{k,m,\tau}
&=\mathbf s^{V,\mathrm{cv}}_{k,m,\tau}
+\Delta\mathbf s^V_{k,m,\tau},\quad \tau=1,\ldots,c.
\label{eq:typed_decoding}
\end{align}
The full skeleton is
\(\hat{\mathbf p}_{i,m,\tau,a}=
\hat{\mathbf p}_{i,m,\tau,0}+
\hat{\boldsymbol\ell}_{i,m,\tau,a}\).
Both endpoints decode from the same pre-chunk \(\mathcal Z_m\), without
consuming each other's new future within the chunk.

Closure, recurrence, and temporal assembly are
\begin{align}
\hat{\mathbf R}^{(m)}
&=\Phi(\mathcal E_m,\hat{\mathbf P}^{(m)},\hat{\mathbf V}^{(m)}),
\nonumber\\
\hat{\mathbf B}^{(m)}
&=\Psi_{\mathrm{box}}(\hat{\mathbf V}^{(m)};
\boldsymbol\xi^{V,\mathrm{obs}}),\nonumber\\
\hat{\mathcal S}^{\mathrm{phys},(m)}
&=(\hat{\mathbf P}^{(m)},\hat{\mathbf V}^{(m)},
\hat{\mathbf R}^{(m)},\hat{\mathbf B}^{(m)}),
\label{eq:chunk_physical_closure}\\
\mathcal E_{m+1}
&=\operatorname{Tail}_2(\hat{\mathbf P}^{(m)},\hat{\mathbf V}^{(m)}),
\quad \mathcal Z_{m+1}=(\mathbf H_{m+1},\mathcal E_{m+1}),
\label{eq:boundary_refresh}\\
\hat{\mathcal S}^{\mathrm{phys}}_{1:T_f}
&=\operatorname{Concat}_{m=0}^{M-1}
\hat{\mathcal S}^{\mathrm{phys},(m)}.
\label{eq:complete_rollout}
\end{align}
For the first generated frame, \(\mathcal E_m\) supplies the preceding
pedestrian root; later relation states use adjacent generated frames.
Final P/V states yield \(\mathbf g^P_{i,m}\) and \(\mathbf g^V_{k,m}\);
the projected final relation and masked temporal pool yield
\(\mathbf g^{R,\mathrm{end}}_{ik,m}\) and
\(\mathbf g^{R,\mathrm{chunk}}_{ik,m}\). These summaries update
\(\mathbf H_{m+1}\).

\subsection{Pair-to-Endpoint Routing Controls}
\label{sec:method_factorization}

Let \(\bar{\mathbf g}^{R,\mathrm{end}\rightarrow X}_{n,m}\) be the
mask-pooled final-relation summary incident to endpoint \(n\) of type
\(X\in\{P,V\}\), and define
\(\Gamma_{\rho_X}(\mathbf x)=\rho_X\mathbf x\).
\(\mathbf c^X_{0,m}\) is within-type, whereas
\(\mathbf c^X_{1,m}\) additionally includes the fused P/V/R summary;
\(\mathbf c^R_m\) is the fused masked P/V/R summary used by the pair
update. The recurrent updates are
\begin{align}
\mathbf h^X_{n,m+1}
&=\mathbf h^X_{n,m}
+U_X\!\left(
\mathbf h^X_{n,m},
\mathbf g^X_{n,m},
\Gamma_{\rho_X}
(\bar{\mathbf g}^{R,\mathrm{end}\rightarrow X}_{n,m}),
\mathbf c^X_{\rho_X,m}
\right),\nonumber\\
\mathbf h^R_{ik,m+1}
&=\mathbf h^R_{ik,m}
+U_R\!\left(
\mathbf h^R_{ik,m},
\mathbf g^{R,\mathrm{chunk}}_{ik,m},
\mathbf c^R_m
\right).
\label{eq:state_update}
\end{align}
Here \(n=i\) for pedestrians and \(n=k\) for vehicles. PV-WM uses
\(\boldsymbol{\rho}=(0,0)\), retaining the pair state for relational
representation and auxiliary readout while excluding pair context from
endpoint initialization, decoding, and update. The matched controls use
\((1,0)\), \((0,1)\), or \((1,1)\), routing observed pair context during
initialization and decoding and generated pair summaries during later
endpoint updates. All settings share the same state, supervision,
horizon, and model budget.

\subsection{Learning Objective}
\label{sec:method_training}

All routing settings minimize
\begin{equation}
\mathcal L=\mathcal L_P+\mathcal L_V+\mathcal L_R+\mathcal L_Q
+0.05\mathcal L_{\mathrm{scene}}.
\label{eq:training_objective}
\end{equation}
\(\mathcal L_P\) uses joint, root, root-relative, body-local, bone,
velocity, and acceleration weights
\((1.0,0.50,0.25,0.10,0.05,0.05,0.02)\);
\(\mathcal L_V\) uses state and center weights \((1.0,0.50)\);
\(\mathcal L_R\) uses R4, distance, and horizon-minimum-distance weights
\((0.35,0.30,0.15)\); and
\(\mathcal L_Q\) uses pairwise and chunk-risk weights \((0.10,0.10)\),
with targets defined as the product of smooth proximity and closing-motion scores. The scene loss regresses observed minimum P--V distance and entity counts. Since relations are
closed from generated endpoints, \(\mathcal L_R\) directly supervises the
P/V predictors. Boxes are reconstructed without a separate regression loss. Future reference states enter training only as supervision targets, while recurrent boundaries and latent updates use generated P/V states.

\section{Experiments}
\label{sec:experiments}

Experiments evaluate forecasting accuracy and complete-system cost,
attribute gains to individual components, and test the recurrent utility
of generated articulation.

\subsection{Experimental Setup}
\label{sec:exp_setup}

\paragraph{Protocol and evaluation}
We align the 15-joint pedestrian tracks of
Waymo-3DSkelMo~\cite{waymo3dskelmo} with vehicle tracks and oriented
boxes from the Waymo Open Dataset~\cite{waymo_open_dataset}, obtaining
824 driving contexts. We extract up to eight \(3\,\mathrm{s}\) windows
per context at a \(1\,\mathrm{s}\) stride and form local scenes as
connected components of an observation-only P--V graph. Vehicles whose
minimum observed distance to any pedestrian is at most
\(25\,\mathrm{m}\) are retained as candidates. A P--V edge is activated
when the observed minimum distance is at most \(12\,\mathrm{m}\), the
minimum time to collision is at most \(4\,\mathrm{s}\), or the maximum
closing speed is at least \(0.25\,\mathrm{m\,s^{-1}}\). Isolated
pedestrians are retained, and each component is capped at eight
pedestrians and eight vehicles, yielding 8,364 local-scene instances.
All methods observe \(20\) frames over \(2\,\mathrm{s}\) and predict
\(10\) frames over \(1\,\mathrm{s}\) at \(10\,\mathrm{Hz}\);
PV-WM executes five two-frame transitions.

We use segment-grouped, context-disjoint five-fold out-of-fold evaluation: contexts, overlapping windows, and tracks from the same Waymo segment remain in the same fold, and every context is held out once. Trainable models use seeds \(42\), \(123\), and \(456\); deterministic CV references are evaluated once. Checkpoints and Modular Specialist compositions are selected using the associated validation split only. For each seed, sufficient statistics are pooled across the five held-out folds before computing the cross-seed mean and sample standard deviation. Paired comparisons use 10,000 context-clustered bootstrap resamples with Holm correction within each prespecified family. Pedestrian metrics use all 824 contexts; vehicle and pairwise metrics
draw from the 797 contexts with valid future vehicle support, with metric-specific masks applied over valid horizons. The main 824-context PV-WM models are map-free and trained from scratch.

\paragraph{Implementation details}
All PV-WM routing settings use 128-dimensional hidden states and are
trained from scratch with Adam, an initial learning rate of
\(10^{-4}\), zero weight decay, and batch size \(6\) for at most
\(30\) epochs. The pedestrian encoder fuses root, body-part,
root-delta, and local-motion MLP streams; vehicle history is encoded by
a one-layer GRU with masked temporal pooling. Chunk heads and recurrent
P/V/R updates use two-layer GELU MLPs with 128-dimensional hidden
layers. Early stopping begins after epoch \(18\), with patience \(6\)
and a minimum relative improvement of \(10^{-3}\). Checkpoints are selected on the corresponding validation split using the equal-weight mean of six errors normalized by the corresponding errors of the CV all-joints + CV-V reference computed on the fold's training split and fixed before
validation: Root ADE, MPJPE, APE, Vehicle ADE, P--V
Distance MAE, and \dcaeobb{}. Each routing setting is trained
independently with the same optimizer, training budget, and
checkpoint-selection rule.

\paragraph{Metrics}
Root ADE and FDE measure pedestrian transport. MPJPE evaluates absolute
15-joint error, while APE removes root translation and evaluates local
articulation. Vehicle ADE and FDE measure planar center error, and
P--V Distance MAE measures framewise
pedestrian-root-to-vehicle-center distance error. Following
distance-of-closest-approach error~\cite{3463952.3464070}, we report
an oriented-box adaptation:
\[
\mathrm{DCAE}_{\mathrm{OBB}}
=
\frac{1}{|\mathcal P_{\mathrm v}|}
\sum_{(i,k)\in\mathcal P_{\mathrm v}}
\left|
\min_{t\in\mathcal T_{ik}}\hat c_{ik,t}
-
\min_{t\in\mathcal T_{ik}}c_{ik,t}
\right|,
\]
where \(\mathcal P_{\mathrm v}\) is the set of valid pedestrian--vehicle pairs, \(\mathcal T_{ik}\) is pair \((i,k)\)'s valid forecast horizon, and \(c_{ik,t}\) is the signed BEV clearance between a \(0.30\,\mathrm{m}\)-radius pedestrian root disk and the reconstructed vehicle OBB. Predicted boxes use generated centers and headings with observation-derived dimensions. \dcaeobb{} measures error in horizon-minimum signed BEV clearance, complementing the framewise P--V Distance MAE.

Cyclic BoxCorner removes arbitrary starting-corner dependence by averaging, over valid vehicle frames, the minimum mean corner distance across the four cyclic shifts of a common clockwise corner ordering~\cite{8578200}. The same matching rule is applied to every method. In the articulation study, fMPJPE denotes final-frame MPJPE. wAPE weights each non-root joint error by the magnitude of its reference root-relative displacement from the last observed pose, clips the weight at \(0.50\,\mathrm{m}\), and normalizes by the total valid weight.

\paragraph{Baselines and complete systems}
Articulated references comprise CV joint extrapolation,
TBIFormer~\cite{tbiformer}, VehCondPose3D~\cite{vehcondpose3d}, and
T2P~\cite{t2p}, paired with the same prescribed CV vehicle rollout.
Trajectron++~\cite{trajectronpp}, QCNet~\cite{qcnet}, and
MTR~\cite{mtr} learn pedestrian-root and vehicle futures but do not
predict articulated pose. Released implementations are adapted to the
common temporal and evaluation protocol; multimodal methods use their
highest-probability mode rather than a test-oracle minimum.

The validation-selected Modular Specialist pairs TBIFormer, T2P, or
VehCondPose3D with vehicle predictions from an independently trained
modular recurrent model, Trajectron++, QCNet, or MTR. The recurrent
model follows the same five two-frame transitions, and only its vehicle
output enters the composition. Each fold selects its composition using the three-seed mean of the same validation score.
\method{} realizes the complete articulated pedestrian--vehicle state
with one end-to-end network. 

\paragraph{System profiling}
Runtime and memory are measured directly on an NVIDIA RTX A5000 in FP32 with batch size one. Each fold uses 50 warm-up executions followed by 200 measured executions. Both Specialist networks are simultaneously resident and executed sequentially. Timing includes in-memory scene decoding, tensor construction and transfer, network execution, P--V geometry assembly, and CUDA synchronization, but excludes disk I/O. The reported p95 pools 1,000 executions across five folds, and memory is the maximum complete-execution peak across folds. FLOPs are measured in a separate pass to avoid profiler overhead and are averaged over all 8,364 local-scene instances using their observed pedestrian and vehicle counts.

\begin{table*}[!t]
\centering
\caption{\textbf{Forecasting accuracy and complete-system cost under the
\(2\,\mathrm{s}\!\rightarrow\!1\,\mathrm{s}\) protocol.}
Results are five-fold OOF mean$\pm$sample standard deviation over three
seeds; the deterministic CV baseline is evaluated once. Errors are in
millimeters. Boldface is assigned within groups with compatible output
spaces; gray text marks methods with prescribed CV vehicle futures.}
\label{tab:main_results}

\small
\setlength{\tabcolsep}{3.2pt}
\renewcommand{\arraystretch}{1.06}

\resizebox{\textwidth}{!}{%
\begin{tabular}{@{}llcccccc@{}}
\toprule
\multirow{2}{*}{\textbf{Method}} &
\multirow{2}{*}{\textbf{Output / Construction}} &
\multicolumn{2}{c}{\textbf{Pedestrian}} &
\textbf{Vehicle} &
\textbf{P--V Geometry} &
\multicolumn{2}{c}{\textbf{System Cost}} \\
\cmidrule(lr){3-4}
\cmidrule(lr){5-5}
\cmidrule(lr){6-6}
\cmidrule(lr){7-8}
&
&
\textbf{Root ADE/FDE (mm)}$\downarrow$ &
\textbf{MPJPE/APE (mm)}$\downarrow$ &
\textbf{ADE/FDE (mm)}$\downarrow$ &
\textbf{Dist./\dcaeobb{} (mm)}$\downarrow$ &
\textbf{N/Params/Avg. FLOPs}$_{\downarrow}$ &
\textbf{p95/Mem. (ms/MiB)}$_{\downarrow}$ \\
\midrule

\multicolumn{8}{@{}l}{
\textit{Articulated pedestrian forecasting with prescribed vehicle motion}
} \\[-1pt]

\gcell{CV all-joints + CV-V} &
\gcell{J15 + prescribed V} &
\gcell{189.1/350.3} &
\gcell{293.3/186.5} &
\gcell{54.6/119.1} &
\gcell{\textbf{110.6/211.1}} &
\gcell{0/\textit{n.r.}/\textit{n.r.}} &
\gcell{\textit{n.r.}} \\

\gcell{TBIFormer~\cite{tbiformer} + CV-V} &
\gcell{J15 + prescribed V} &
\gcell{184.0\sdev{10.5}/316.8\sdev{21.0}} &
\gcell{216.2\sdev{9.3}/87.2\sdev{0.1}} &
\gcell{54.6/119.1} &
\gcell{132.8\sdev{8.1}/230.0\sdev{7.2}} &
\gcell{1/\textit{n.r.}/\textit{n.r.}} &
\gcell{\textit{n.r.}} \\

\gcell{VehCondPose3D~\cite{vehcondpose3d} + CV-V} &
\gcell{J15 + prescribed V} &
\gcell{\textbf{176.0}\sdev{5.2}/\textbf{304.3}\sdev{10.7}} &
\gcell{\textbf{204.6}\sdev{3.1}/82.5\sdev{1.6}} &
\gcell{54.6/119.1} &
\gcell{128.3\sdev{4.3}/225.2\sdev{3.8}} &
\gcell{1/\textit{n.r.}/\textit{n.r.}} &
\gcell{\textit{n.r.}} \\

\gcell{T2P~\cite{t2p} + CV-V} &
\gcell{J15 + prescribed V} &
\gcell{389.3\sdev{39.8}/702.9\sdev{76.2}} &
\gcell{406.2\sdev{39.1}/\textbf{79.7}\sdev{1.9}} &
\gcell{54.6/119.1} &
\gcell{289.2\sdev{40.7}/362.9\sdev{37.7}} &
\gcell{1/\textit{n.r.}/\textit{n.r.}} &
\gcell{\textit{n.r.}} \\

\midrule
\multicolumn{8}{@{}l}{
\textit{Trajectory-native learned pedestrian--vehicle forecasting}
} \\[-1pt]

Trajectron++~\cite{trajectronpp} &
Root + learned V &
179.1\sdev{0.8}/326.8\sdev{1.1} &
\NA &
\textbf{64.8}\sdev{1.2}/138.7\sdev{5.2} &
\textbf{103.0}\sdev{1.2}/421.8\sdev{7.0} &
1/\textit{n.r.}/\textit{n.r.} &
\textit{n.r.} \\

QCNet~\cite{qcnet} &
Root + learned V &
\textbf{143.6}\sdev{1.3}/\textbf{245.1}\sdev{2.5} &
\NA &
68.9\sdev{4.4}/\textbf{128.9}\sdev{6.0} &
111.7\sdev{2.8}/\textbf{361.0}\sdev{2.4} &
1/\textit{n.r.}/\textit{n.r.} &
\textit{n.r.} \\

MTR~\cite{mtr} &
Root + learned V &
146.5\sdev{1.5}/251.6\sdev{2.4} &
\NA &
83.5\sdev{5.8}/155.0\sdev{9.8} &
116.2\sdev{2.7}/382.4\sdev{2.4} &
1/\textit{n.r.}/\textit{n.r.} &
\textit{n.r.} \\

\midrule
\multicolumn{8}{@{}l}{
\textit{Complete articulated pedestrian--vehicle systems}
} \\[-1pt]

\rowcolor{gray!6}
Modular Specialist$^\dagger$ &
Val.-selected \(3P\times4V\) &
174.2\sdev{4.0}/300.0\sdev{7.8} &
204.3\sdev{1.9}/84.0\sdev{1.6} &
\textbf{68.1}\sdev{0.9}/\textbf{146.8}\sdev{2.4} &
128.4\sdev{3.3}/198.2\sdev{2.0} &
2/8.757M/2.534G &
29.46/201.09 \\

\rowcolor{gray!16}
\textbf{\method{}} &
\textbf{J15 + learned V + OBB + R} &
\textbf{165.2}\sdev{2.3}/\textbf{288.4}\sdev{6.9} &
\textbf{188.8}\sdev{1.7}/\textbf{83.9}\sdev{2.2} &
68.9\sdev{0.4}/148.2\sdev{1.6} &
\textbf{113.1}\sdev{0.2}/\textbf{186.6}\sdev{0.8} &
\textbf{1/3.756M/89.71M} &
\textbf{21.94/65.59} \\

\bottomrule
\end{tabular}%
}

\vspace{0.3mm}
\begin{minipage}{0.995\textwidth}
\scriptsize
J15 denotes a predicted 15-joint skeleton; for PV-WM, OBBs are reconstructed and \(R\) is recomputed from generated P/V states; N is the number of learned networks, and \textit{n.r.} denotes unavailable matched profiling.
\(^{\dagger}\)The Modular Specialist is selected independently per fold from \(3P\times4V\) candidates using the six-metric validation composite (fold 0: TBIFormer + modular recurrent vehicle branch; folds 1--4: VehCondPose3D + modular recurrent vehicle branch); held-out data are never used for selection.
Runtime and memory are direct complete-system measurements; FLOPs are averaged over 8,364 local-scene instances.
Pedestrian metrics use 824 contexts; vehicle/P--V metrics draw from
the 797 contexts with valid future support and apply metric-specific
masks.
\end{minipage}

\end{table*}

\begin{figure*}[!t]
\centering
\includegraphics[width=\textwidth]{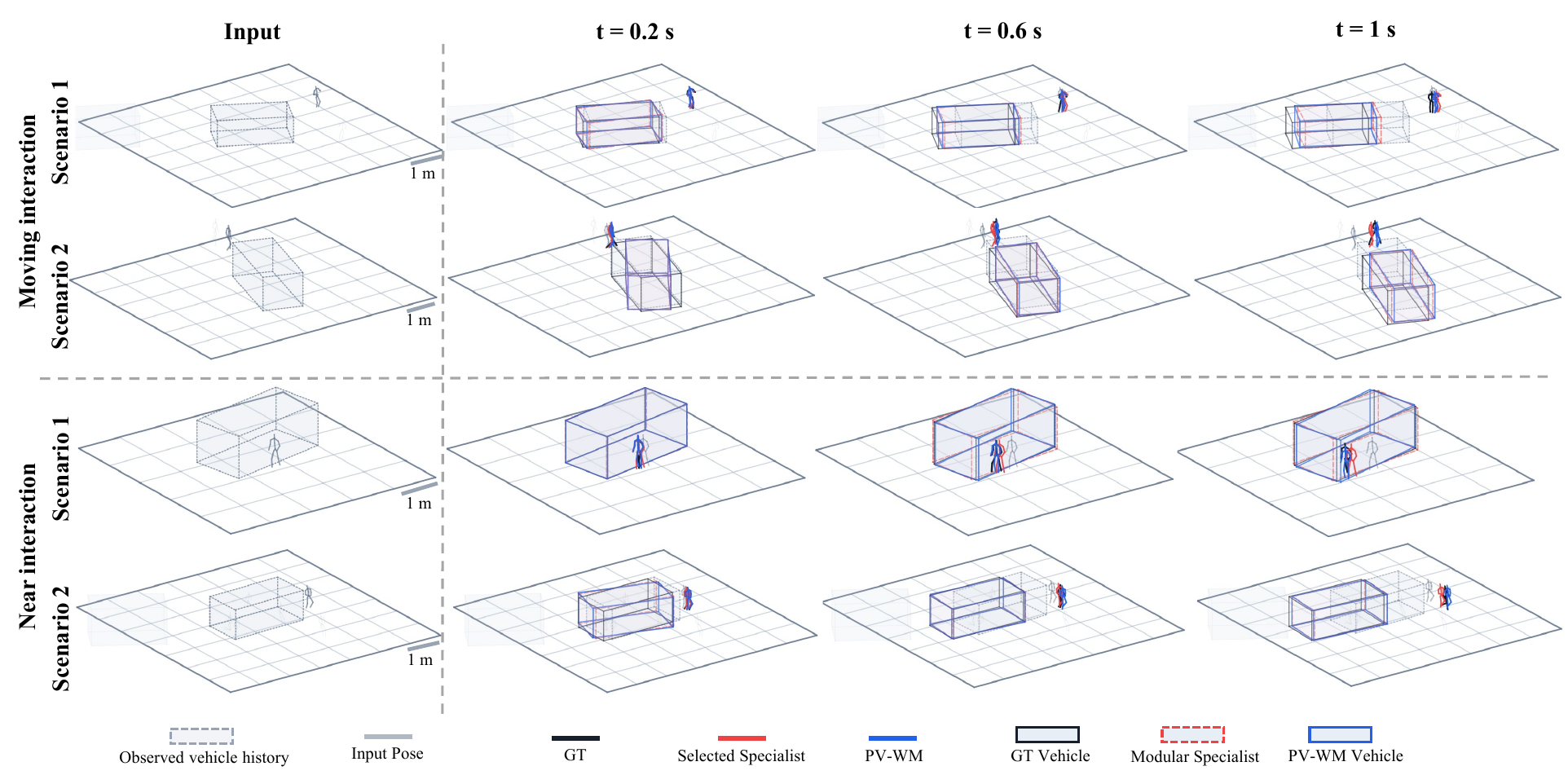}
\caption{\textbf{Complete-state forecasts on held-out contexts.}
Rows contain two moving and two near-interaction contexts; columns show the observation and predictions at \(0.2\), \(0.6\), and \(1.0\,\mathrm{s}\). Black denotes the reference, red the Modular Specialist, and blue \method{}; skeletons and vehicle OBBs jointly visualize articulated motion, rigid-body motion, and P--V geometry.}
\label{fig:qualitative_complete_state}
\end{figure*}

\subsection{Forecasting Accuracy and Complete-System Cost}
\label{sec:exp_main}

Table~\ref{tab:main_results} groups methods by the future state they
predict. Among articulated methods, \method{} obtains the lowest Root
ADE/FDE and MPJPE while learning vehicle futures; T2P has lower APE, and
trajectory-native methods achieve lower root error but omit articulation.

Among complete-state systems, \method{} improves Root ADE, MPJPE, P--V
Distance MAE, and \dcaeobb{} by \(9.00\), \(15.51\), \(15.31\), and
\(11.56\,\mathrm{mm}\), or \(5.2\%\), \(7.6\%\), \(11.9\%\), and
\(5.8\%\), relative to the Modular Specialist
(\(p_{\mathrm{Holm}}\leq0.0012\)). APE is statistically indistinguishable,
whereas Vehicle ADE increases by \(0.85\,\mathrm{mm}\).

Using one rather than two networks, \method{} reduces the parameter
count by \(57.1\%\), average FLOPs per local scene by \(96.5\%\), p95
latency by \(25.5\%\), and peak memory by \(67.4\%\).

\paragraph{Qualitative results.}
Figure~\ref{fig:qualitative_complete_state} shows close agreement in
body configuration and relative P--V geometry in the first three
contexts; the final near-interaction case exhibits late-horizon root
and box displacement.

\subsection{Component and Extension Ablations}
\label{sec:exp_ablation}

Table~\ref{tab:architecture_ablation} isolates recurrent execution,
typed endpoint dynamics, persistent pair state, relational supervision,
and pair-to-endpoint routing under generated-state rollout. 

\begin{table*}[!t]
\centering
\caption{\textbf{PV-WM component and extension ablations.}
The main block uses generated-state continuation; static-context
extensions share the same 339 map-present contexts and are ranked
separately. One-shot prediction is non-recurrent and generated-context
training is training-only. The dark row is the
Table~\ref{tab:main_results} configuration. Errors are in millimeters;
bold marks the lowest value within each support block.}
\label{tab:architecture_ablation}

\scriptsize
\setlength{\tabcolsep}{1.45pt}
\renewcommand{\arraystretch}{0.98}

\begin{adjustbox}{max width=\textwidth,center}
\begin{tabular}{@{}l cccccccccc ccccc@{}}
\toprule
\multirow[c]{2}{*}{\textbf{Variant}}
& \multicolumn{10}{c}{\textbf{Design}}
& \multicolumn{2}{c}{\textbf{Pedestrian}}
& \multicolumn{2}{c}{\textbf{Vehicle}}
& \textbf{P--V Geometry} \\
\cmidrule(lr){2-11}
\cmidrule(lr){12-13}
\cmidrule(lr){14-15}
\cmidrule(lr){16-16}
& \textbf{Rec.}
& \textbf{Typed}
& \textbf{Pair}
& \textbf{Geo.}
& \textbf{Q}
& \textbf{\(R\!\rightarrow\!P\)}
& \textbf{\(R\!\rightarrow\!V\)}
& \textbf{Gen. Ctx.}
& \textbf{Map}
& \textbf{Rule}
& \textbf{Root ADE/FDE (mm)}$_{\downarrow}$
& \textbf{MPJPE/APE (mm)}$_{\downarrow}$
& \textbf{ADE/FDE (mm)}$_{\downarrow}$
& \textbf{Cyclic BoxCorner (mm)}$_{\downarrow}$
& \textbf{Dist./\dcaeobb{} (mm)}$_{\downarrow}$\\
\midrule

\multicolumn{16}{@{}l}{
\textit{Reference configuration}
} \\[-1pt]

\rowcolor{gray!16}
\textbf{\method{}}
& \cmark & \cmark & G & \cmark & \cmark
& \xmark & \xmark & \xmark & \xmark & \xmark
& \textbf{\abpm{165.2}{2.3}/\abpm{288.4}{6.9}}
& \textbf{\abpm{188.8}{1.7}}/\abpm{83.9}{2.2}
& \abpm{68.9}{0.4}/\abpm{148.2}{1.6}
& \abpm{329.4}{1.8}
& \abpm{113.1}{0.2}/\textbf{\abpm{186.6}{0.8}} \\

\midrule
\multicolumn{16}{@{}l}{
\textit{Core component ablations}
} \\[-1pt]

One-shot complete-state prediction
& \xmark & \cmark & O & \cmark & \cmark
& \xmark & \xmark & \xmark & \xmark & \xmark
& \abpm{189.3}{0.0}/\abpm{350.6}{0.1}
& \abpm{221.6}{1.5}/\abpm{83.2}{1.6}
& \textbf{\abpm{57.3}{0.4}/\abpm{123.3}{0.4}}
& \abpm{430.7}{11.4}
& \textbf{\abpm{110.7}{0.0}}/\abpm{214.9}{2.8} \\

Shared endpoint update
& \cmark & \xmark & G & \cmark & \cmark
& \xmark & \xmark & \xmark & \xmark & \xmark
& \abpm{186.9}{1.4}/\abpm{343.4}{4.9}
& \abpm{206.1}{3.6}/\abpm{91.6}{3.8}
& \abpm{65.5}{0.6}/\abpm{141.9}{1.6}
& \abpm{349.9}{6.5}
& \abpm{111.6}{0.2}/\abpm{189.4}{2.7} \\

w/o persistent pair state
& \cmark & \cmark & -- & \cmark & \cmark
& \xmark & \xmark & \xmark & \xmark & \xmark
& \abpm{188.1}{0.2}/\abpm{347.4}{0.8}
& \abpm{208.7}{2.0}/\abpm{91.0}{3.3}
& \abpm{67.4}{0.6}/\abpm{145.6}{1.7}
& \abpm{377.8}{8.2}
& \abpm{112.1}{0.1}/\abpm{195.9}{0.8} \\

w/o explicit geometry losses
& \cmark & \cmark & G & \xmark & \cmark
& \xmark & \xmark & \xmark & \xmark & \xmark
& \abpm{168.7}{3.7}/\abpm{298.8}{10.3}
& \abpm{191.4}{3.1}/\abpm{85.6}{2.1}
& \abpm{70.7}{0.7}/\abpm{149.1}{1.4}
& \textbf{\abpm{328.8}{7.0}}
& \abpm{113.2}{0.3}/\abpm{187.9}{0.4} \\

w/o Q supervision
& \cmark & \cmark & G & \cmark & \xmark
& \xmark & \xmark & \xmark & \xmark & \xmark
& \abpm{171.6}{3.2}/\abpm{304.7}{7.0}
& \abpm{192.9}{1.8}/\abpm{87.3}{2.2}
& \abpm{67.7}{0.7}/\abpm{145.5}{0.7}
& \abpm{336.3}{3.2}
& \abpm{112.0}{0.3}/\abpm{187.6}{0.3} \\

w/o all relational supervision
& \cmark & \cmark & G & \xmark & \xmark
& \xmark & \xmark & \xmark & \xmark & \xmark
& \abpm{166.8}{1.3}/\abpm{293.7}{4.7}
& \abpm{190.2}{0.9}/\abpm{84.7}{1.5}
& \abpm{70.4}{0.7}/\abpm{147.9}{1.2}
& \abpm{336.1}{7.8}
& \abpm{113.4}{0.1}/\abpm{190.5}{1.8} \\

\midrule
\multicolumn{16}{@{}l}{
\textit{Pair-to-endpoint routing controls}
} \\[-1pt]

+\(R\!\rightarrow\!P\) routing
& \cmark & \cmark & G & \cmark & \cmark
& \cmark & \xmark & \xmark & \xmark & \xmark
& \abpm{190.5}{0.6}/\abpm{353.6}{1.0}
& \abpm{216.7}{1.6}/\abpm{84.2}{3.0}
& \abpm{65.3}{0.2}/\abpm{140.2}{1.4}
& \abpm{368.1}{10.9}
& \abpm{111.6}{0.0}/\abpm{196.5}{1.8} \\

+\(R\!\rightarrow\!V\) routing
& \cmark & \cmark & G & \cmark & \cmark
& \xmark & \cmark & \xmark & \xmark & \xmark
& \abpm{168.2}{4.1}/\abpm{295.0}{9.5}
& \abpm{191.3}{3.2}/\abpm{84.5}{2.5}
& \abpm{69.3}{1.2}/\abpm{148.8}{3.2}
& \abpm{373.6}{11.6}
& \abpm{112.7}{0.4}/\abpm{194.8}{1.6} \\

+\(R\!\rightarrow\!\{P,V\}\) routing
& \cmark & \cmark & G & \cmark & \cmark
& \cmark & \cmark & \xmark & \xmark & \xmark
& \abpm{190.5}{0.4}/\abpm{353.7}{1.4}
& \abpm{217.4}{1.3}/\abpm{83.4}{1.6}
& \abpm{66.1}{0.8}/\abpm{141.3}{2.6}
& \abpm{383.6}{16.6}
& \abpm{111.5}{0.1}/\abpm{200.1}{1.9} \\

\midrule
\multicolumn{16}{@{}l}{
\textit{Training extension}
} \\[-1pt]

\method{} + generated-context training
& \cmark & \cmark & G & \cmark & \cmark
& \xmark & \xmark & \cmark & \xmark & \xmark
& \abpm{165.4}{2.9}/\abpm{289.3}{7.0}
& \abpm{189.4}{3.2}/\textbf{\abpm{82.4}{0.5}}
& \abpm{68.6}{0.3}/\abpm{147.9}{0.9}
& \abpm{346.1}{14.3}
& \abpm{112.9}{0.2}/\abpm{191.6}{3.0} \\

\midrule
\multicolumn{16}{@{}l}{
\textit{Static-context extensions on 339 common-support contexts}
} \\[-1pt]

\rowcolor{gray!4}
\method{} on map support
& \cmark & \cmark & G & \cmark & \cmark
& \xmark & \xmark & \xmark & \xmark & \xmark
& \textbf{\abpm{165.3}{2.1}/\abpm{292.5}{6.1}}
& \textbf{\abpm{190.4}{1.7}/\abpm{85.9}{1.8}}
& \textbf{\abpm{79.2}{0.3}/\abpm{175.7}{1.7}}
& \textbf{\abpm{358.2}{4.5}}
& \textbf{\abpm{114.6}{0.4}}/\textbf{\abpm{206.9}{1.2}} \\

\method{} + MapCtx
& \cmark & \cmark & G & \cmark & \cmark
& \xmark & \xmark & \xmark & \cmark & \xmark
& \abpm{170.0}{2.2}/\abpm{305.4}{5.2}
& \abpm{193.3}{1.8}/\abpm{88.0}{1.2}
& \abpm{79.6}{1.6}/\abpm{176.1}{2.7}
& \abpm{370.4}{5.3}
& \textbf{\abpm{114.6}{0.5}}/\abpm{210.4}{0.7} \\

\method{} + MapCtx + SoftRule
& \cmark & \cmark & G & \cmark & \cmark
& \xmark & \xmark & \xmark & \cmark & \cmark
& \abpm{176.1}{1.0}/\abpm{320.3}{3.9}
& \abpm{197.8}{0.5}/\abpm{89.4}{1.0}
& \abpm{112.4}{2.0}/\abpm{499.7}{8.8}
& \abpm{406.7}{8.1}
& \abpm{139.7}{1.0}/\abpm{222.7}{3.8} \\

\bottomrule
\end{tabular}
\end{adjustbox}

\vspace{0.3mm}
\begin{minipage}{0.995\textwidth}
\scriptsize
Pair is observation-fixed (O), generated (G), or removed (--).
Geo. combines R4, distance-channel, and horizon-minimum-distance
losses; Q is auxiliary risk supervision. Arrows indicate
pair-to-endpoint routing. Gen. Ctx. denotes generated-context training,
a second pass that appends a detached two-frame generated prefix to the
temporally shifted observation history before predicting the remaining
horizon. MapCtx adds gated static context and SoftRule its proxy
objective. Pedestrian metrics use 824 contexts; vehicle/P--V metrics draw from
797 contexts with valid future support and apply metric-specific masks;
map rows use 339 contexts and are ranked separately.
\end{minipage}
\end{table*}

\begin{table*}[!t]
\centering
\caption{\textbf{Utility and content specificity of articulated state.}
Panel (a) separates articulated prediction from recurrent feedback;
bold marks the lowest displayed error and the gain row reports Rigid
minus Stateful. Panel (b) reports final-horizon changes under feedback
interventions and oracle feedback. All errors are in millimeters.}
\label{tab:articulation_utility}

\scriptsize
\renewcommand{\arraystretch}{1.08}

\noindent
\begin{minipage}[t]{0.72\textwidth}
\vspace{0pt}
\centering
\setlength{\tabcolsep}{1.8pt}

\resizebox{\linewidth}{!}{%
\begin{tabular}{@{}l cc ccc c c@{}}
\toprule
\multicolumn{8}{c}{
\textit{(a) Representation and recurrent-state utility}
} \\
\midrule

\multirow[c]{2}{*}{\textbf{Variant}}
& \multicolumn{2}{c}{\textbf{Articulated State}}
& \multicolumn{3}{c}{\textbf{Pedestrian}}
& \textbf{Vehicle}
& \textbf{P--V Geometry} \\
\cmidrule(lr){2-3}
\cmidrule(lr){4-6}
\cmidrule(lr){7-7}
\cmidrule(lr){8-8}

& \textbf{Readout}
& \textbf{Feedback}
& \textbf{Root ADE/FDE}$_{\downarrow}$
& \textbf{MPJPE/fMPJPE}$_{\downarrow}$
& \textbf{APE/wAPE}$_{\downarrow}$
& \textbf{ADE}$_{\downarrow}$
& \textbf{Dist./\dcaeobb{}}$_{\downarrow}$ \\
\midrule

\rowcolor{gray!5}
Rigid state
& \xmark
& \xmark
& 185.17\sdev{0.92}/337.98\sdev{3.62}
& 218.98\sdev{0.88}/364.59\sdev{3.48}
& 84.91\sdev{0.00}/228.40\sdev{0.00}
& \textbf{66.29\sdev{0.28}}
& \textbf{112.09\sdev{0.24}}/188.25\sdev{1.04} \\

Articulated readout
& \cmark
& \xmark
& 182.11\sdev{5.25}/329.13\sdev{14.60}
& 211.32\sdev{5.76}/354.42\sdev{14.43}
& 78.94\sdev{1.17}/200.84\sdev{5.18}
& 66.68\sdev{1.44}
& 112.23\sdev{0.53}/188.31\sdev{3.17} \\

\rowcolor{gray!16}
\textbf{Stateful articulation}
& \cmark
& \cmark
& \textbf{179.15\sdev{3.64}/321.30\sdev{8.42}}
& \textbf{207.86\sdev{3.57}/346.19\sdev{8.12}}
& \textbf{77.25\sdev{0.14}/185.35\sdev{0.38}}
& 67.76\sdev{0.57}
& 112.61\sdev{0.09}/\textbf{186.82\sdev{1.95}} \\

\midrule
\multicolumn{3}{@{}l}{
\textit{Gain: Rigid \( - \) Stateful}
}
& \textbf{+6.02/+16.68}
& \textbf{+11.13/+18.40}
& \textbf{+7.67/+43.05}
& \(-1.48\)
& \(-0.52/+1.42^{\mathrm{n.s.}}\) \\

\bottomrule
\end{tabular}%
}
\end{minipage}
\hfill
\begin{minipage}[t]{0.26\textwidth}
\vspace{0pt}
\centering
\setlength{\tabcolsep}{2.2pt}

\resizebox{\linewidth}{!}{%
\begin{tabular}{@{}l l cc@{}}
\toprule
\multicolumn{4}{c}{
\textit{(b) Feedback content specificity}
} \\
\midrule

\multirow[c]{2}{*}{\textbf{Intervention}}
& \multirow[c]{2}{*}{\textbf{Tests}}
& \multicolumn{2}{c}{
\textbf{Final-horizon error change (mm)}
} \\
\cmidrule(lr){3-4}
&
& \textbf{\(\Delta\)fMPJPE}
& \textbf{\(\Delta\)wAPE@1s} \\
\midrule

Rigidized
& Content
& +2.79
& +33.94 \\

Time shuffle
& Time
& +2.72
& +33.66 \\

Identity shuffle$^\dagger$
& Identity
& +33.63
& +32.54 \\

\midrule

Oracle GT
& Headroom
& -14.06
& -101.51 \\

\bottomrule
\end{tabular}%
}
\end{minipage}

\vspace{0.3mm}
\begin{minipage}{0.995\textwidth}
\scriptsize
fMPJPE is final-frame MPJPE; wAPE weights root-relative error by
future-target body motion. GT denotes the reconstructed future target.
For Stateful versus Rigid in panel (a), the P--V Distance MAE increase
is Holm-significant; \(^{\mathrm{n.s.}}\) marks the nonsignificant
\dcaeobb{} contrast. Panel (b) reports changes from factual feedback; positive values indicate
degradation and negative values oracle headroom. Identity shuffling
applies to 717 eligible multi-pedestrian contexts; the others retain
factual feedback, yielding the same 824-context aggregation. Other
interventions apply to all 824 contexts. First-chunk predictions are identical, and all displayed contrasts in panel (b) are Holm-significant.
\end{minipage}
\end{table*}

\paragraph{Recurrent and typed execution.}
Relative to the matched one-shot control, \method{} lowers Root ADE,
MPJPE, \dcaeobb{}, and Cyclic BoxCorner by \(24.1\), \(32.8\),
\(28.3\), and \(101.3\,\mathrm{mm}\), respectively. The one-shot
control yields lower APE, Vehicle ADE/FDE, and framewise P--V Distance
MAE. Replacing typed endpoint updates with a shared transition increases
Root ADE, MPJPE, \dcaeobb{}, and Cyclic BoxCorner by \(21.7\),
\(17.3\), \(2.8\), and \(20.5\,\mathrm{mm}\); removing the persistent
pair state increases them by \(22.9\), \(19.9\), \(9.3\), and
\(48.4\,\mathrm{mm}\).

\paragraph{Relational learning and routing}
Removing the full relational objective increases \dcaeobb{} by
\(3.9\,\mathrm{mm}\) after Holm correction. Removing only explicit
geometry losses changes \dcaeobb{} by \(+1.3\,\mathrm{mm}\) and
Cyclic BoxCorner by \(-0.6\,\mathrm{mm}\); only the full relational
objective yields a Holm-significant \dcaeobb{} effect. Routing pair
context to both endpoint branches increases Root ADE, MPJPE, Cyclic
BoxCorner, and \dcaeobb{} by \(25.3\), \(28.6\), \(54.2\), and
\(13.5\,\mathrm{mm}\), while reducing APE, Vehicle ADE, and framewise
P--V Distance MAE.

Taken together, these controls separate three roles: generated
boundaries close the recurrent loop, typed updates preserve
heterogeneous dynamics, and the persistent pair state carries
relational context while the endpoint branches remain factorized.

\paragraph{Training and context extensions}
Generated-context training trades small local-pose and vehicle-center
gains for \(16.7\,\mathrm{mm}\) higher Cyclic BoxCorner and
\(5.0\,\mathrm{mm}\) higher \dcaeobb{}; PV-WM therefore uses the
base training objective. On the 339 map-present contexts, neither
MapCtx nor SoftRule improves the map-free reference.

\subsection{Utility of Articulated Pedestrian State}
\label{sec:exp_articulation}

We separate articulated output, structured supervision, and recurrent
state reuse under the default factorized endpoint dynamics of PV-WM. \emph{Rigid State} predicts root
motion while repeating the final observed root-relative pose.
\emph{Articulated Readout} predicts and supervises future pose without
feeding it into later chunks. \emph{Stateful Articulation} additionally
re-encodes generated pose into the next recurrent boundary. This
independently trained suite uses the same folds and seeds, a fixed
30-epoch budget, and a common checkpoint criterion that excludes pose
metrics; comparisons are therefore made within this study.

On fixed Stateful Articulation checkpoints, interventions modify root-relative pose feedback at each chunk boundary before latent updates and boundary refresh. Rigidization repeats the last observed local pose; time shuffling swaps local poses between the two chunk frames; identity shuffling cyclically permutes local poses among valid pedestrians within each local scene; oracle feedback substitutes reference local poses. Each edit preserves the predicted roots and vehicle states at that boundary and leaves the decoded chunk unchanged.

Stateful Articulation reduces Root ADE and FDE by \(6.02\) and
\(16.68\,\mathrm{mm}\), MPJPE and final-frame MPJPE by \(11.13\) and
\(18.40\,\mathrm{mm}\), and APE and motion-weighted APE by
\(7.67\) and \(43.05\,\mathrm{mm}\) relative to Rigid State. All six
effects have the same direction across the five folds and remain
significant after Holm correction. Vehicle ADE and P--V Distance MAE increase by \(1.48\) and
\(0.52\,\mathrm{mm}\), respectively; the \dcaeobb{} difference is not significant.

Articulated Readout reduces MPJPE by \(7.66\,\mathrm{mm}\) and wAPE by
\(27.56\,\mathrm{mm}\) relative to Rigid State; recurrent feedback
provides a further \(3.46\,\mathrm{mm}\) MPJPE and
\(15.49\,\mathrm{mm}\) wAPE improvement. Rigidizing or time-shuffling
the feedback increases final-frame MPJPE by \(2.79\) and
\(2.72\,\mathrm{mm}\), and endpoint wAPE by \(33.94\) and
\(33.66\,\mathrm{mm}\). Identity shuffling increases final-frame
MPJPE by \(33.63\,\mathrm{mm}\), while oracle feedback retains
\(14.06\,\mathrm{mm}\) fMPJPE and \(101.51\,\mathrm{mm}\) wAPE@1s
headroom. Because first-chunk predictions are identical, later transitions depend
on the content, temporal order, and identity of generated articulation.

\section{Conclusion}
\label{sec:conclusion}

We introduced \method{}, a heterogeneous micro--macro world model that
recurrently advances articulated pedestrian and learned vehicle states,
reconstructs oriented boxes, and synchronizes P--V geometry after each
transition. Ablations identify recurrent typed execution and the
persistent pair branch as the main architectural contributors, while
relational supervision improves closest-approach accuracy. Feedback
interventions show that subsequent transitions functionally reuse
generated articulation.

On the segment-grouped Waymo evaluation, single-network \method{}
improves pedestrian transport, full-body accuracy, and P--V geometry
over a validation-selected Modular Specialist while reducing parameters
by \(57.1\%\) and average FLOPs per local scene by \(96.5\%\).
PV-WM demonstrates that a recurrent world model can jointly advance
articulated pedestrian and rigid-vehicle states while preserving
type-specific dynamics and synchronized geometry.

\section*{ACKNOWLEDGMENT}
The schematic local-scene illustration in the upper-left of
Fig.~\ref{fig:method} was generated with the image-generation
capability of OpenAI ChatGPT and subsequently edited by the authors;
it contains no experimental data.

\bibliographystyle{IEEEtran}
\bibliography{references}

\end{document}